\documentclass[times,twocolumn,final]{elsarticle}

\usepackage{framed,multirow}

\usepackage{amssymb}
\usepackage{latexsym}

\usepackage{subfigure}
\usepackage{epstopdf}
\usepackage{array} 
\usepackage{booktabs}

\usepackage{algorithm}
\usepackage{algorithmic}

\usepackage{url}
\usepackage{xcolor}
\definecolor{newcolor}{rgb}{.8,.349,.1}

\begin{document}

\begin{frontmatter}

\title{Toward high-performance online HCCR: a CNN approach with \emph{DropDistortion}, path signature and spatial stochastic max-pooling}

\author[1]{Songxuan Lai}
\author[1]{Lianwen Jin}
\author[1]{Weixin Yang}

\address[1]{School of Electronic and Information Engineering, South China University of Technology, Guangzhou, China}

\begin{abstract}
This paper presents an investigation of several techniques that increase the accuracy of online handwritten Chinese character recognition (HCCR). We propose a new training strategy named \emph{DropDistortion} to train a deep convolutional neural network (DCNN) with distorted samples. \emph{DropDistortion} gradually lowers the degree of character distortion during training, which allows the DCNN to better generalize. Path signature is used to extract effective features for online characters. Further improvement is achieved by employing spatial stochastic max-pooling as a method of feature map distortion and model averaging. Experiments were carried out on three publicly available datasets, namely CASIA-OLHWDB 1.0, CASIA-OLHWDB 1.1, and the ICDAR2013 online HCCR competition dataset. The proposed techniques yield state-of-the-art recognition accuracies of 97.67\%, 97.30\%, and 97.99\%, respectively.
\\
\end{abstract}


\end{frontmatter}


\section{Introduction}
Deep convolutional neural networks (DCNNs) have brought about breakthroughs in many domains, and their application to online handwritten Chinese character recognition (HCCR) has persistently yielded state-of-the-art results in recent years \citep{DBLP:journals/corr/Graham13,DBLP:journals/corr/Graham14a,Yang2015Improved,yang2016dropsample}. These previous work have effectively improved HCCR performance, e.g. application of the path signature theory \citep{DBLP:journals/corr/Graham13}, usage of comprehensive domain knowledge \citep{Yang2015Improved} and more advanced training methods \citep{yang2016dropsample}, etc. However, several challenges still need to be addressed.

The main challenge presented by HCCR results from varied handwriting styles and the large number of Chinese character classes. A large-scale dataset is vital to HCCR performance, especially when using a DCNN. However, data acquisition with high quality ground-truth is a tedious job. One possible solution to this problem may be to apply character distortion to generate artificial samples \citep{DBLP:journals/corr/Graham13,Yang2015Improved,JIN2002A,Leung2009Recognition}. Such kind of approach enables the generation of a large number of training samples to improve the performance. However, in previous studies, character distortion is usually applied at a fixed degree throughout the entire training process. Even though a fixed high-degree distortion reduces overfitting by generating varied samples, it may lead to a distribution that deviates from the underlying data distribution, whereas a fixed low-degree distortion would achieve the opposite. Hence, more proper distortion strategies should be investigated.

Another challenge is to find a good feature representation for online characters. Although DCNN is good at capturing visual concepts from raw inputs, prior knowledge can be encoded into the inputs for a DCNN to improve the performance\citep{Yang2015Improved}. In online HCCR, 8-directional features \citep{bai2005study}, or path signature \citep{2016SignaturePrimer} can be regarded as prior knowledge that enhances a DCNN, but it needs further study to determine how these feature representations could be improved. Rather than incorporating dynamic information into DCNN, \citep{DBLP:journals/corr/ZhangYZLB16} used recurrent neural network (RNN) to directly recognize and draw online Chinese characters, and showed that RNN is also able to learn complex representation of online characters.

In this paper, we focus on DCNN based models and explore several techniques to address these challenges. Motivated by the great performance of the path signature in HCCR \citep{DBLP:journals/corr/Graham13}, we utilize path signature as features for online characters. Different from previous usage of path signature features for HCCR, we add a time dimension to the online characters in accordance with the writing order to enable extraction of more expressive features. Our DCNN architecture is carefully designed to enable more effective learning from the signature features, with spatial stochastic max-pooling layers performing feature map distortion and model averaging. The \emph{DropDistortion} training strategy, which gradually lowers the character distortion degree during training, is proposed to address the drawbacks of a fixed degree of distortion. Fig. 1 pictorially illustrates this concept. In early training epochs, a high degree of distortion provides improved generalization, whereas in later epochs, a decreasing degree of distortion gradually reveals more genuine data distribution. Experiments on the CASIA-OLHWDB 1.0, CASIA-OLHWDB 1.1, and the ICDAR2013 online HCCR competition dataset achieve state-of-the-art accuracies of 97.67\%, 97.30\% and 97.99\%, respectively.

\begin{figure}[t]
  \centering
  \includegraphics[]{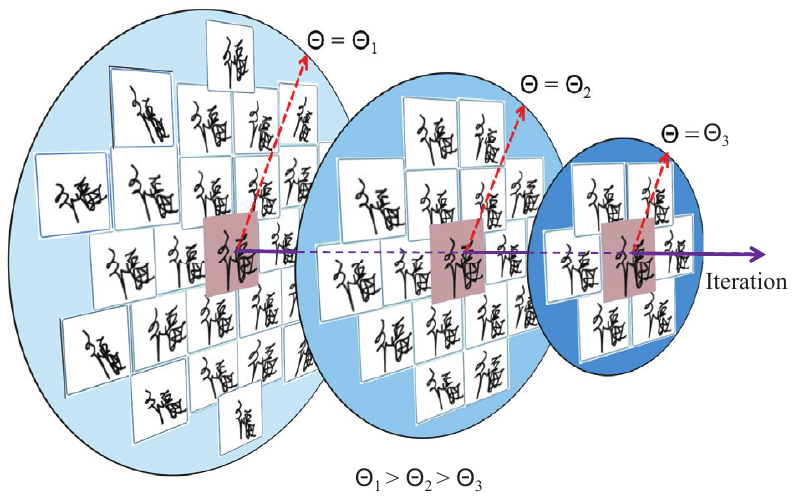}
  \caption{Illustration of \emph{DropDistortion}. The distortion degree $\Theta$ is lowered after certain epochs at the training stage, in order to gradually reveal the genuine data distribution.}\label{figureDrop}
\end{figure}

The remaining part of this paper is organized as follows. Section \uppercase\expandafter{\romannumeral2} provides a detailed analysis of the proposed \emph{DropDistortion} method. Section \uppercase\expandafter{\romannumeral3} introduces the path signature. Section \uppercase\expandafter{\romannumeral4} describes our DCNN architecture and design pipeline. Section \uppercase\expandafter{\romannumeral5} presents the experimental results and its detailed analysis. Finally, Section \uppercase\expandafter{\romannumeral6} concludes the paper.

\section{\emph{DropDistortion} training strategy}
Character distortion is widely used in HCCR to generate artificial training samples\citep{DBLP:journals/corr/Graham13,Yang2015Improved}. The basic structures of the distorted samples are the same as those of the originals. Fig. 2 illustrates some rotated samples of a Chinese character. In previous studies, however, character distortion is applied upto a certain extent in an empirical way. In this paper, we study the influence of character distortion in detail and propose a simple but novel method, namely \emph{DropDistortion}, to help enhance HCCR performance.

\begin{figure}[t]
  \centering
  \includegraphics{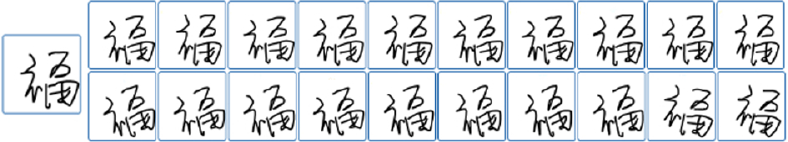}
  \caption{Rotational distortion. The leftmost one is the original Chinese character. The first and second rows correspond to lower and higher degrees of rotational distortion, respectively.}\label{figureRotation}
\end{figure}

\subsection{Character distortion via affine transformation}
We use affine transformation to distort input characters. Let $\Theta$ denote the degree of character distortion, \emph{$\xi$} denote a random number drawn from uniform distribution \emph{U(-$\Theta$, $\Theta$)}, and [\emph{\textbf{x}} \emph{\textbf{y}}] denote the coordinate series of a character stroke, a matrix of size \emph{n}$\times$2 where n indicates the number of data points. Then we can apply affine transformation to distort online handwritten Chinese characters:
\begin{equation}\label{equation1}
  [\emph{\textbf{x}}\ \emph{\textbf{y}}]\Leftarrow[\emph{\textbf{x}}\ \emph{\textbf{y}}]\cdot\left[
                                                                                       \begin{array}{cc}
                                                                                         1+\emph{$\xi_x$} & 0 \\
                                                                                         0 & 1+\emph{$\xi_y$} \\
                                                                                       \end{array}
                                                                                     \right],
\end{equation}
\begin{equation}\label{equation2}
  [\emph{\textbf{x}}\ \emph{\textbf{y}}]\Leftarrow[\emph{\textbf{x}}\ \emph{\textbf{y}}]\cdot\left[
                                                                                       \begin{array}{cc}
                                                                                         1 & \emph{$\xi$} \\
                                                                                         0 & 1 \\
                                                                                       \end{array}
                                                                                     \right],
\end{equation}
\begin{equation}\label{equation3}
  [\emph{\textbf{x}}\ \emph{\textbf{y}}]\Leftarrow[\emph{\textbf{x}}\ \emph{\textbf{y}}]\cdot\left[
                                                                                       \begin{array}{cc}
                                                                                         1 & 0 \\
                                                                                         \emph{$\xi$} & 1 \\
                                                                                       \end{array}
                                                                                     \right],
\end{equation}
\begin{equation}\label{equation4}
  [\emph{\textbf{x}}\ \emph{\textbf{y}}]\Leftarrow[\emph{\textbf{x}}\ \emph{\textbf{y}}]\cdot\left[
                                                                                       \begin{array}{cc}
                                                                                         \cos(\emph{$\xi$}) & -\sin(\emph{$\xi$}) \\
                                                                                         \sin(\emph{$\xi$}) & \cos(\emph{$\xi$}) \\
                                                                                       \end{array}
                                                                                     \right],
\end{equation}
where (1) stretches or shrinks the stroke, (2) and (3) slant the stroke and (4) performs rotational distortion. A character is distorted by simply applying one or more of the above equations to all its strokes, with \emph{$\xi$} in the same equation fixed within a character. We also randomly translate the characters to achieve a better distortion diversity.

\subsection{Analysis of character distortion from the Bayesian perspective}
First, consider a simple example with two handwritten characters, the Arabic numeral ``1'' and the Chinese character``one'' (Fig. 3). We use \emph{A} and \emph{B} to denote them respectively for convenience. When written by hand \emph{A} and \emph{B} look quite similar except for their orientations. In other words, they have the same topological structure. Let \emph{$\theta$} denote the angle between their orientations and the horizontal direction, and assume \emph{$\theta$} obeys a Gaussian distribution: \emph{$\theta_A$}$\sim$\emph{N($\pi$/2,$\sigma^2$)}, \emph{$\theta_B$}$\sim$\emph{N(0,$\sigma^2$)}.

\begin{figure}[t]
  \centering
  \subfigure[Arabic 1]{
    \label{subfig:num1} 
    \includegraphics{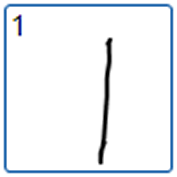}}
  \hspace{2ex}
  \subfigure[one]{
    \label{subfig:one} 
    \includegraphics{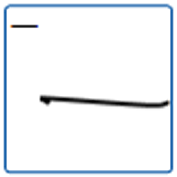}}
  \hspace{2ex}
  \subfigure[the sun]{
    \label{subfig:the sun} 
    \includegraphics{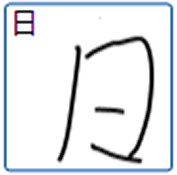}}
  \hspace{2ex}
  \subfigure[say]{
    \label{subfig:say} 
    \includegraphics{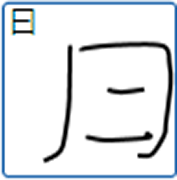}}
  \caption{Easily confused Characters when distorted.}
  \label{figureCharacters} 
\end{figure}

Then from the Bayesian perspective, we have
\begin{equation}\label{equation5}
  f(A|\theta)=f(\theta|A)f(A)/f(\theta),
\end{equation}
\begin{equation}\label{equation6}
  f(B|\theta)=f(\theta|B)f(B)/f(\theta),
\end{equation}
where \emph{f(A$|$$\theta$)} and \emph{f(B$|$$\theta$)} are probabilities predicted by a classifier, \emph{f($\theta$$|$A)} and \emph{f($\theta$$|$B)} model the orientation information, \emph{f(A)} and \emph{f(B)} model the topological structure information, and \emph{f($\theta$)} is a normalized constant determined by the training set and can be safely removed:
\begin{equation}\label{equation7}
  f(A|\theta)\propto f(\theta|A)f(A),
\end{equation}
\begin{equation}\label{equation8}
  f(B|\theta)\propto f(\theta|B)f(B).
\end{equation}
If the topological structure of \emph{A} and \emph{B} is nearly the same, we have
\begin{equation}\label{equation9}
  f(A)\approx f(B).
\end{equation}
Although \emph{f(A)}$\approx$\emph{f(B)}, the classifier is still able to correctly classify these two characters according to \emph{$\theta$} due to the difference between \emph{f($\theta$$|$A)} and \emph{f($\theta$$|$B)}.

\begin{figure*}[tb]
  \centering
  \subfigure[$\Theta$=$\pi$/6]{
    \label{subfig:a} 
    \includegraphics{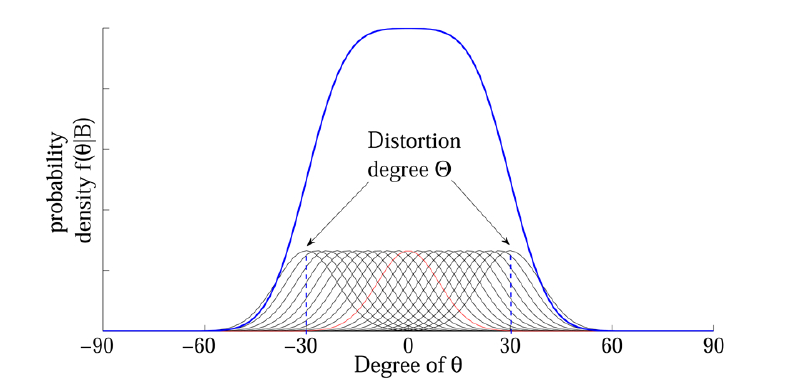}}
  \subfigure[$\Theta$=$\pi$/3]{
    \label{subfig:b} 
    \includegraphics{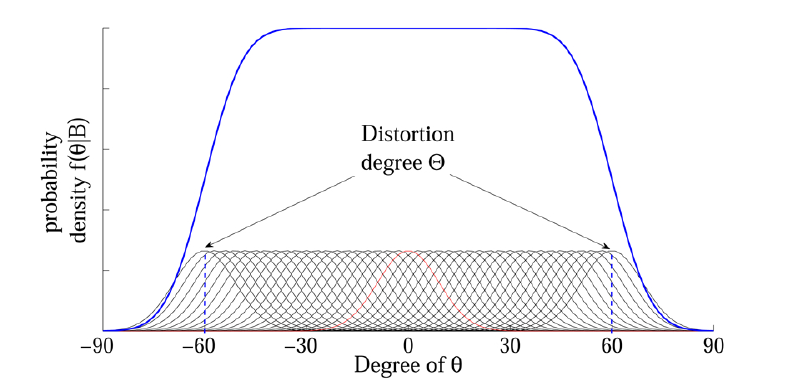}}
  \caption{Conditional distribution of $\theta$ given \emph{B}, a mixture of Gaussians. \emph{f($\theta$$|$A)} is exactly the same case.}
  \label{figureGaussian} 
\end{figure*}

If rotational distortion is applied to \emph{A} and \emph{B} with $\xi$ drawn from the uniform distribution \emph{U(-$\Theta$, $\Theta$)}, then the distribution of \emph{$\theta$} is shown in Fig. 4. If rotational distortion is applied to an extreme, i.e., $\Theta$=$2\pi$, then \emph{$\theta$} obeys a uniform distribution as well. Under these circumstances, the classifier is unable to distinguish between \emph{A} and \emph{B} because \emph{f($\theta$$|$A)}=\emph{f($\theta$$|$B)} and \emph{f(A)}$\approx$\emph{f(B)}.

However, when it comes to HCCR, the situation is somewhat different. A rotated Chinese character will rarely resemble other Chinese characters. A high degree of rotational distortion actually removes the conditional term \emph{f($\theta$$|$C)} and forces the classifier to predict \emph{f(C)} more accurately, i.e., to achieve improved learning of the topological structure of the character \emph{C}. By preventing co-adaptation of \emph{f(C)} and \emph{f($\theta$$|$C)}, a high degree of rotational distortion reduces overfitting. Other distortions such as stretching can be analyzed in the same way. Hence using character distortion to generate artificial samples is a reasonable and effective way to enhance HCCR performance.

\subsection{DropDistortion: stepwise character distortion}
Although effective in improving HCCR performance, character distortion changes the data distribution. Specially, it confuses some similar characters inevitably. For example, the Chinese characters ``the sun'' (Fig. 3 (c)) and ``say'' (Fig. 3 (d)) would be indistinguishable if they were to be highly stretched. The change of distribution should be considered to give further improvements, but no such efforts have been reported in previous studies \citep{DBLP:journals/corr/Graham13,Yang2015Improved,JIN2002A,Leung2009Recognition}, where the character distortion is carried out upto a fixed degree during the entire training process. The proposed \emph{DropDistortion} method is a novel strategy that is designed to take the change of distribution into consideration. It's based on a simple idea that the DCNN should be fine-tuned with more genuine samples, i.e., low-degree or non distorted samples. The proposed \emph{DropDistortion} algorithm is given in algorithm 1.

\begin{algorithm}[t]
\caption{\small \emph{DropDistortion}}
\textbf{Input:} training set $X$={($x_i$, $y_i$), i=1,...,m of k classes}.\\
\textbf{Initialize:} index $n\leftarrow1$; distortion degree $\Theta_1>\Theta_2>...>\Theta_N$.\\
\textbf{Output:} DCNN parameters $W$. \\
\textbf{Begin:}
\begin{algorithmic}
\WHILE{$n<N$}
\STATE $\Theta\leftarrow\Theta_n$
\WHILE{not converge}
  \STATE Sample distortion at degree $\Theta_n$
  \STATE Update W through back propagation algorithm
\ENDWHILE
\STATE $n\leftarrow n+1$
\ENDWHILE
\end{algorithmic}
\label{alg_lirnn}
\end{algorithm}

In the proposed method, a high degree of distortion is used in the early training epochs to generate varied samples to help the DCNN learn effective features and reduce the risk of overfitting. In the later epochs, the degree of distortion decreases gradually to allow a subsequent finer adjustment of the DCNN with more genuine samples. The only extra complexity \emph{DropDistortion} introduces is to monitor the training loss and decrease the distortion rate if a certain condition is fulfilled. In practice, \emph{DropDistortion} can be simply implemented in a multi-step way, and in this paper it is implemented in a three-step way as explained in Fig. 1.

\section{Representation of online handwritten characters with path signature}
In mathematics, a path signature is a collection of iterated integrals \citep{2016SignaturePrimer,Chen1957Integration} of a path. 
Let \emph{X}:[0, \emph{T}]$\rightarrow$\emph{$\mathbb{R}^d$} denote a continuous path of bounded variation, mapping from time interval [0, \emph{T}] to space \emph{$\mathbb{R}^d$}. Then the \emph{k}-th iterated integral of path \emph{X} is
\begin{equation}\label{pathIntegral}
  I^k=\int_{0<t_1<...<t_k<T}1dX_{t_1}\otimes\ldots\otimes dX_{t_k},
\end{equation}
where $\otimes$ represents the tensor product. By convention, $I^0$ is the number one. The signature of path \emph{X} is the collection of all the iterated integrals of \emph{X}, denoted by \emph{S(X)}. Since this is an infinite series, in practice one often considers the first \emph{m}th-order integrals, namely truncated path signature:
\begin{equation}\label{signature}
  S(X)|_m=\{1, I^1, I^2, ..., I^m\}.
\end{equation}
As is often the case, \emph{X} is sampled and approximated by a set of discrete points. Then the iterated integrals can be approximated by using some simple tensor algebra \citep{DBLP:journals/corr/Graham13}.

Online handwriting can be seen as a path mapping from time interval [0, \emph{T}] to $\mathbb{R}^2$. Graham \citep{DBLP:journals/corr/Graham13} first introduced truncated path signature features to online handwritten character recognition as inputs for a DCNN and achieved remarkable performance. Some previous work \citep{DBLP:journals/corr/Graham13,yang2016deepwriterid,Yang2015Chinese} experimented with truncated level \emph{m} and found that the accuracy increases with the increase of \emph{m} and gradually saturates. In our work, \emph{m} is empirically set as 4 because integrals of a higher order typically characterize more trivial details of a path and do not lead to further improvement.

The signature is invariant to translations and time re-parameterization of the path, and uniquely characterizes the path if it contains no part that exactly retraces itself \citep{Hambly2005Uniqueness,Boedihardjo2014The}. However, in handwritten characters, some local parts do retrace themselves due to joined-up writing. For example, path ([0,0], [1.5,2.5], [3,3], [1.5,2.5]) has the same signature with path ([0,0], [1.5,2.5]) because ([1.5,2.5], [3,3], [1.5,2.5]) exactly retraces itself. To handle this problem, we introduce a monotone time dimension to the original handwriting sequences, i.e., the \emph{n}$\times$2 matrix [\emph{\textbf{x}} \emph{\textbf{y}}] is appended by a column vector \emph{\textbf{t}}=$[0\ 1\ 2\ ...\ n-1]^T$. This concept is demonstrated in Fig. 5. The time dimension ensures the uniqueness for the path signature, hence features extracted from the \emph{n}$\times$3 matrix [ \emph{\textbf{t}} \emph{\textbf{x}} \emph{\textbf{y}}] would be more expressive.

In practice, if \emph{m}=4, then \emph{N}=31 for a 2D character and \emph{N}=121 for a 3D character, where \emph{N} denotes the number of input channels.

\begin{figure}[t]
  \centering
  \includegraphics[width=3in]{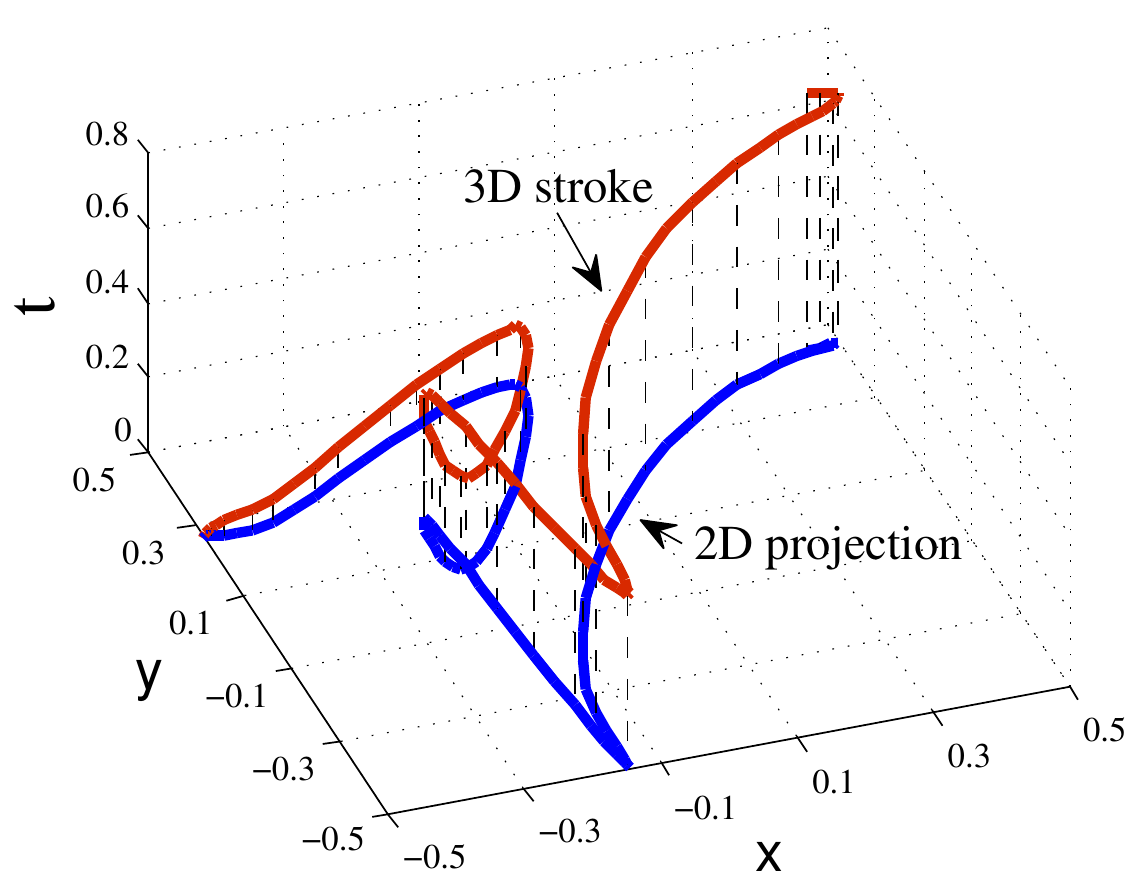}
  \caption{Time dimension is added as a third dimension. The 2D projection is the original stroke.}\label{figure5}
\end{figure}

\section{Deep convolutional neural network with spatial stochastic max-pooling}
Deep convolutional neural networks (DCNNs) have shown great success in computer vision and pattern recognition, and different architectures of DCNN have been explored \citep{Krizhevsky2012ImageNet,DBLP:journals/corr/LinCY13,Szegedy2015Going,DBLP:journals/corr/HeZRS15}. However, these designs did not evaluate the max-pooling (MP) operation, which incorporates a degree of invariance with respect to translations and elastic distortions into the DCNN \citep{lecun1998gradient}. Max-pooling operates in a sliding window method, and conventionally the pooling region slides with a fixed integer stride. Usually the stride is two, hence the size of feature map is reduced by a factor of two.
\subsection{Spatial stochastic max-pooling}
Compared to traditional MP layers, fractional max-pooling (FMP) \citep{DBLP:journals/corr/Graham14a} layers reduce the feature map size by a factor of $\alpha$ with $\alpha$ as a fraction. If 1$<$$\alpha$$<$2, then FMP reduces the size in a slower manner than MP. As $\alpha$ is a fraction, the pooling stride cannot be a fixed integer. In this case, the pooling region slides with a pre-calculated stride series. We prefer to call FMP as spatial stochastic max-pooling (SSMP), which describes the pooling process more precisely.

Let \emph{N$_{in}$} and \emph{N$_{out}$} denote the input and output feature map size respectively. Then
\begin{equation}\label{equationFMP}
  N_{out}=\lfloor N_{in}/\alpha+0.5\rfloor,
\end{equation}
\begin{equation}\label{equationRenew}
  \alpha\leftarrow N_{in}/N_{out}.
\end{equation}
$\alpha$ is renewed above due to the rounding effect. The stride series is a $1\times N_{out}$ vector for each dimension of the output image. Then for the i-th (i=0,1,...$N_{out}-$2) position of the stride series,
\begin{equation}\label{equationRenew}
  a_i=\lfloor i\times\alpha+th_i\rfloor,
\end{equation}
\begin{equation}\label{equationRenew}
  s_i=a_i-a_{i-1}.
\end{equation}
where $s_i$ is the stride series, $a_i$ is the accumulated stride series with $a_{-1}=0$ and $a_{N_{out}-1}=N_{in}-2$ (the pooling size is 2$\times$2), and $th_i\in[0,1)$ is a randomly drawn threshold to round $a_i$ up or down. In practice, $th_i$ can be set independently at different position i, or set only once and shared across different positions as a constant. We denote these two strategies as SSMP$_1$ and SSMP$_2$ respectively. Different to \citep{DBLP:journals/corr/Graham14a} which employs SSMP$_1$, we propose a new strategy that $th_i$ is set independently at each position i in early training epochs and set only once in the last epochs. We call this strategy as SSMP$_3$.

During the pooling process, SSMP introduces a certain degree of randomness into the pooling regions by a random choice of the feasible stride series. Fig. 6 illustrates this concept vividly. The colored 2$\times$2 squares are the pooling regions chosen. By setting $\alpha$ as 1.5, the feature map size is reduced from 6$\times$6 to 4$\times$4 with multiple feasible pooling choices; hence every forward path may have a slightly different output. Moreover, SSMP achieves elastic distortion \citep{DBLP:journals/corr/Graham14a} of the feature maps of the previous layer, which implicitly distorts the input characters and improves the generalization ability of a DCNN. In test phase, running the network for multiple times and averaging the outputs achieve the same effect as an ensemble of similar networks.
\begin{figure}[t]
  \centering
  \subfigure{\label{fmp1}\includegraphics{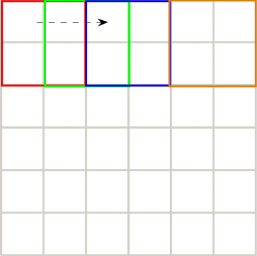}}
  \subfigure{\label{fmp2}\includegraphics{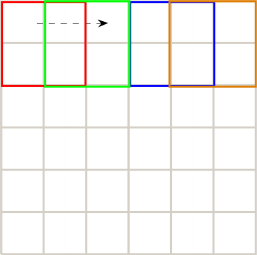}}
  \subfigure{\label{fmp3}\includegraphics{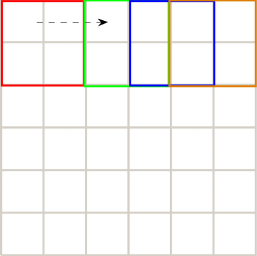}}
  \caption{Random pooling regions of SSMP(1.5). Pooling stride series from left to right: 112, 121, 211.}\label{figure6}
\end{figure}

\subsection{Other details}
In our DCNN, we use 3$\times$3 or 2$\times$2 convolutional filters \citep{DBLP:journals/corr/Graham13}, which also makes the network deeper compared to larger filter size.
The path signature features are extracted from online handwriting data sequences, and then padded into \emph{N}$\times$50$\times$50 bitmaps embedded in larger grids (pad zeros around the characters).
The first pooling layer is max-pooling rather than SSMP, as we find max-pooling works better than SSMP when dealing with the input, partly due to the input noise introduced by padding a continuous handwriting path onto a discrete bitmap.
We apply the leaky ReLU activation function \citep{Andrew2013LeakyReLU,DBLP:journals/corr/XuWCL15} with \emph{a}=0.333, which outperforms ReLU \citep{Vinod2010Rectified} in terms of convergence speed and representation capability \citep{DBLP:journals/corr/XuWCL15}.
\section{Experiments}
\subsection{Dataset}
The datasets we used are CASIA-OLHWDB 1.0 (DB 1.0), CASIA-OLHWDB 1.1 (DB 1.1) \citep{Liu2011CASIA} and ICDAR2013 competition dataset \citep{yin2013icdar}. DB 1.0 contains 3740 Chinese character classes in standard level-1 set of GB2312-80 (GB1) and is obtained from 420 writers (336$\times$3740 samples for training, 84$\times$3740 for testing). DB 1.1 contains 3755 classes in GB1 and is obtained from 300 writers (240$\times$3755 samples for training, 60$\times$3755 for testing). The database for the ICDAR2013 online HCCR competition consists of three datasets containing isolated Chinese characters, namely, CASIA-OLHWDB 1.0$\sim$1.2 (DB 1.0$\sim$1.2). The test set was released after the competition, which contains 3755 classes in GB1.  

\subsection{Network training}
In the experiments, we use nesterov momentum with momentum $\mu$=0.9. The learning rate is set as 0.003 initially, and halved after the first iteration, and then exponentially decreases to 0.00001. The training mini-batch size is 96. We trained the DCNN for 70 epochs.

\subsection{Experimental results}
\begin{figure}[tb]
  \centering
  \includegraphics{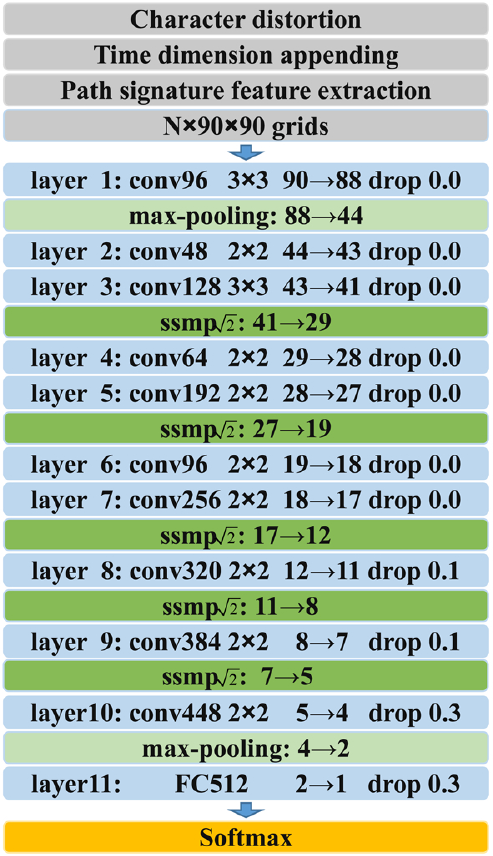}
  \caption{Our proposed DCNN architecture. The ``$\rightarrow$'' denotes the reduction of feature map size. The value following ``drop" means the dropout ratio.}\label{figure7}
\end{figure}
\subsubsection{Individual evaluation of the proposed methods}
We first conducted experiment on DB1.1 to evaluate the effectiveness of our proposed method. i.e, \emph{DropDistortion}, SSMP$_3$ and 3D signature. We designed a baseline DCNN with simple 0-1 bitmaps as input, and fixed the distortion rate $\Theta=0.3$. We evaluated the three techniques individually through the variable-controlling approach. The architecture of the baseline DCNN is 
Input - 32C3 - MP2 - 64C3 - 96C3 - MP2 - 128C3 - 160C3 - MP2 - 192C3 - 224C3 - MP2 - 256C3 - Output.

In order to evaluate the \emph{DropDistortion}, the baseline DCNN is trained with \emph{DropDistortion}.

In order to evaluate the 3D signature features, the baseline DCNN is trained with 2D signature and 3D signature.

For the evaluation of the SSMP and the $th_i$ drawing methods as described in section 4.1, i.e. SSMP$_1$, SSMP$_2$ and SSMP$_3$, two MP layers in the baseline DCNN is replaced with four SSMP layers with $\alpha$=1.5. Refer to Table 1. The last convolutional layer is 256C2 instead of 256C3 because $1.5^4>2^2$. In the test phase, repeatedly running the network produce a slightly different output each time because of SSMP. Averaging these outputs can improve the accuracy.

\begin{table*}[tb]
\renewcommand{\arraystretch}{1.0}
\caption{Configuration of SSMP DCNN by replacing MP layers with SSMP layers.}
\label{tableNetwork}
\centering
\footnotesize
\begin{tabular}{m{2.0cm}<{\centering} m{0.4cm}<{\centering} m{0.4cm}<{\centering} m{0.4cm}<{\centering} m{0.4cm}<{\centering} m{0.8cm}<{\centering} m{0.4cm}<{\centering} m{0.8cm}<{\centering} m{0.4cm}<{\centering} m{0.8cm}<{\centering} m{0.4cm}<{\centering} m{0.8cm}<{\centering}  m{0.4cm}<{\centering}  m{0.4cm}<{\centering}  m{0.4cm}<{\centering}}
\toprule
dimension&32&--&64&96&--&128&--&160&--&192&--&224&--&256\\
\midrule
Baseline DCNN & C3 & MP2 & C3 & C3 & \textbf{MP2} & C3 & () & C3 & \textbf{MP2} & C3 & () & C3 & \textbf{MP2} & C3\\
SSMP DCNN     & C3 & MP2 & C3 & C3 & \textbf{SSMP1.5} & C3 & \textbf{SSMP1.5} & C3 & \textbf{SSMP1.5} & C3 & \textbf{SSMP1.5} & C3 & \textbf{MP2} & C2\\
\bottomrule
\end{tabular}
\end{table*}

The experimental results are presented in Table 2. The employed three techniques, namely \emph{DropDistortion}, 3D signature and SSMP$_3$, all improve the performance over the baseline and their corresponding counterparts. The 2D signature substantially improves the performance, while the 3D signature gives further improvement. SSMP greatly decreases the error rates by averaging 10 test scores, and the SSMP$_3$ we proposed is better than SSMP$_1$ and SSMP$_2$. \emph{DropDistortion} decreases the error rate with little extra cost. As \emph{DropDistortion} and SSMP can be used jointly with path signature, we conducted further experiments to evaluate the joint effects.
\begin{table}[tb]
\renewcommand{\arraystretch}{1.0}
\caption{Evaluation of the proposed three techniques: \emph{DropDistortion}, SSMP and path signature (test error rates, \%)}
\label{tableEva}
\centering
\begin{tabular}{c c c } 
\toprule
Models&1 test&10 tests\\ 
\midrule
Baseline($\Theta=0.3$)&4.70&---\\
\emph{DropDistortion}&4.55&---\\
2D signature&3.18&---\\
3D signature&3.08&---\\
SSMP$_1$&4.78&4.30\\
SSMP$_2$&4.75&4.25\\
SSMP$_3$&4.68&4.23\\
\bottomrule
\end{tabular}
\end{table}

\subsubsection{Joint evaluation of the proposed methods}
To achieve state-of-art results, we extended the baseline model to be deeper and wider. The overall architecture is shown in Fig. 7. The second (48C2), fourth (64C2) and sixth layer (96C2) reduce the dimension in lower layers and act as regularization. The experimental results are presented in Table 3, where we can see \emph{DropDistortion} always achieves the best performance, and a higher degree distortion is usually better than a lower degree distortion. \emph{DropDistortion} brings about an obvious improvement when using bitmaps as inputs, as bitmaps only contains structural information and \emph{DropDistortion} helps to learn the structure.

Compared to rendering online characters as bitmaps, the path signature greatly improves the recognition accuracy, which proves its effectiveness in information extraction. And the 3D signature also gives consistent improvement over the 2D signature.

The SSMP averaging results are given in Table 4 (\emph{DropDistortion} + 3D signature + SSMP$_3$). By averaging 10 test scores of the network, the test error rates decrease significantly on both DB1.1 and ICDAR2013 competition dataset.

The best results for DB1.1 and ICDAR2013 competition dataset are both produced by jointly using \emph{DropDistortion}, 3D signature and SSMP, which results in relative error reduction of 36.3\% (=(4.24-2.70)/4.24*100\%) and 37.2\% (=(3.20-2.01)/3.20*100\%)\, respectively.

\begin{table*}[tbph]
\renewcommand{\arraystretch}{1.0}
\caption{Test error rates(\%) on CASIA-OLHWDB 1.1 and ICDAR2013 online HCCR competition dataset}
\label{table1}
\centering
\begin{tabular}{m{2cm}<{\centering} m{2.4cm}<{\centering} m{2.5cm}<{\centering} m{2.5cm}<{\centering} m{2.5cm}<{\centering}}\hline
\toprule
&&\multicolumn{3}{c}{Character Distortion $\Theta$}\\\cline{3-5}
\multicolumn{1}{c}{\raisebox{1.5ex}[0pt]{Dataset}}&\multicolumn{1}{c}{\raisebox{1.5ex}[0pt]{Features}}&fixed 0.1&fixed 0.3&DropDistortion: 0.3$\rightarrow$0.2$\rightarrow$0.1\\
\midrule
&Bitmap&4.24&4.18&4.08\\
DB 1.1&2D signature&3.10&3.02&2.99\\
&3D signature&3.01&2.95&2.92\\
\midrule
&Bitmap&3.20&3.25&3.13\\
ICDAR2013&2D signature&2.32&2.27&2.24\\
&3D signature&2.30&2.26&2.21\\
\bottomrule
\end{tabular}
\end{table*}

\begin{table*}[tb]
\renewcommand{\arraystretch}{1.0}
\caption{Test error rates(\%) of SSMP model averaging}
\label{table2}
\centering
\begin{tabular}{m{3cm}<{\centering} m{1cm}<{\centering} m{1cm}<{\centering} m{1cm}<{\centering} m{1cm}<{\centering} m{1cm}<{\centering} m{1cm}<{\centering} m{1cm}<{\centering} m{1cm}<{\centering} m{1cm}<{\centering} m{1.2cm}<{\centering}}
\toprule
Database&1 test&2 tests&3 tests&4 tests&5 tests&6 tests&7 tests&8 tests&9 tests&10 tests\\
\midrule
DB 1.1 & 2.92 & 2.82 & 2.78 & 2.75 & 2.75 & 2.73 & 2.71 & 2.70 & 2.70 & 2.70 \\
ICDAR2013 & 2.21 & 2.11 & 2.09 & 2.08 & 2.05 & 2.05 & 2.03 & 2.03 & 2.03 & 2.01 \\
\bottomrule
\end{tabular}
\end{table*}

\subsubsection{Comparison with published state-of-the-art results}
We conducted further experiments on DB 1.0, and in Table 5 we compared our method with some state-of-the-art achieved for HCCR in previous studies. It can be seen from Table 5 that our method has achieved the highest recognition accuracies for all the three datasets, showing the effectiveness of the proposed approach.

Our DCNN architecture in Fig. 7 follows that of DeepCNet and FMP network, e.g. 2$\times$2 convolutional kernels, increasing kernel numbers and stacked SSMP layers, but it is no deeper than the network in \citep{DBLP:journals/corr/Graham14a}. Domain knowledge and \emph{DropSample} can be adapted into our approach to give further improvement, but this is beyond the scope of this paper. 8-DirectMap contains directional information, which is equivalent to the first level of path signature\citep{yang2016dropsample}. Higher levels of path signature we used here contain richer information and can further improve the performance.

Fig. 8 demonstrates some randomly chosen misclassified samples, from which we can discern that the misclassified samples are usually illegible. Some are even mislabeled or wrongly written.

\begin{figure}[tb]
  \centering
  \includegraphics{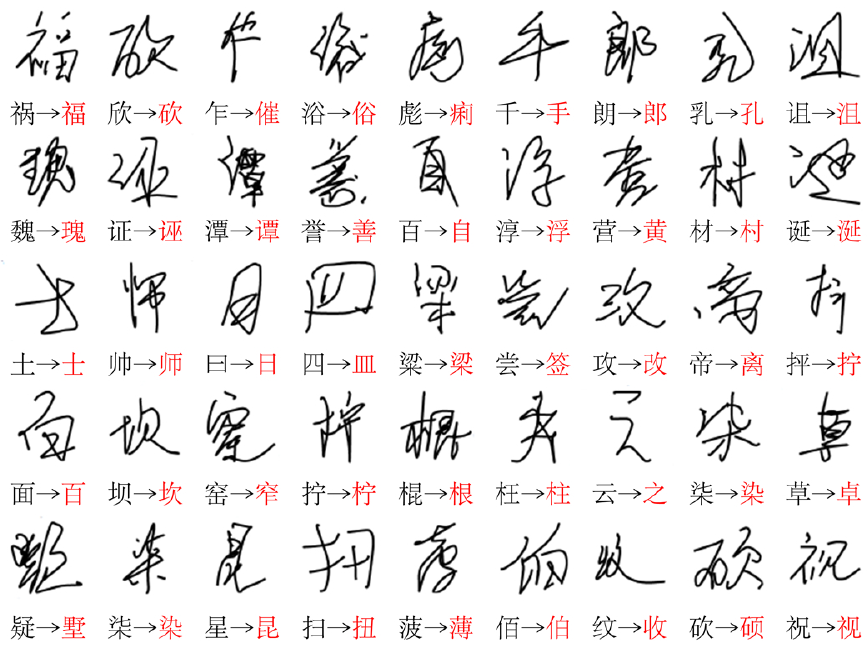}
  \caption{Examples of some misclassified samples. The printed Chinese characters under the samples are the ground truth class (left) and the predicted class (right).}
  \label{misclassified}
\end{figure}

\begin{table}[tb]
\renewcommand{\arraystretch}{1.0}
\caption{Comparison with published state-of-the-art accuracies(\%)}
\label{table3}
\centering
\begin{tabular}{m{3cm}<{\centering} c c m{2cm}<{\centering}} 
\toprule
Methods&DB 1.0&DB 1.1&ICDAR2013 Competition\\ 
\midrule
DeepCNet&&&\\\citep{DBLP:journals/corr/Graham13}&\raisebox{1.0ex}[0pt]{---}&\raisebox{1.0ex}[0pt]{96.42}&\raisebox{1.0ex}[0pt]{97.39}\\
Domain Knowledge&&&\\\citep{Yang2015Improved}&\raisebox{1.0ex}[0pt]{97.20}&\raisebox{1.0ex}[0pt]{96.87}&\raisebox{1.0ex}[0pt]{97.20}\\
FMP ensemble&&&\\\citep{DBLP:journals/corr/Graham14a}&\raisebox{1.0ex}[0pt]{---}&\raisebox{1.0ex}[0pt]{97.03}&\raisebox{1.0ex}[0pt]{---}\\
DropSample&&&\\\citep{yang2016dropsample}&\raisebox{1.0ex}[0pt]{97.33}&\raisebox{1.0ex}[0pt]{97.06}&\raisebox{1.0ex}[0pt]{97.51}\\
DirectMap+convnet&&&\\\citep{Zhang2016Online}&\raisebox{1.0ex}[0pt]{---}&\raisebox{1.0ex}[0pt]{---}&\raisebox{1.0ex}[0pt]{97.55}\\
\midrule
Our Method 1 test&97.5&97.08&97.79\\
&&&\\
\raisebox{1ex}[0pt]{Our Method 10 tests}&\raisebox{1.0ex}[0pt]{\textbf{97.67}}&\raisebox{1.0ex}[0pt]{\textbf{97.30}}&\raisebox{1.0ex}[0pt]{\textbf{97.99}}\\
\bottomrule
\end{tabular}
\end{table}

\section{Conclusion}
This paper presents several new techniques for online HCCR that effectively boost the recognition accuracy. We proposed a simple but effective character distortion method called \emph{DropDistortion}, which improves the recognition accuracy with little additional computational cost. Path signature acts as an effective feature representation for online characters, and the SSMP layers in our DCNN perform feature map distortion and model averaging. Experiments on CASIA-OLHWDB 1.0, CASIA-OLHWDB 1.1 and the ICDAR2013 online HCCR competition dataset achieved new state-of-the-art accuracies of 97.67\%, 97.30\% and 97.99\%, respectively. Compared with previous best result \citep{Zhang2016Online} on the ICDAR2013 competition dataset, our method has achieved an error reduction of 17.9\%, showing the effectiveness of the proposed approach.

Although we mainly focus on online HCCR, the proposed \emph{DropDistortion} method is expected to serve as a general technique in many other tasks in machine learning, such as image classification. The \emph{DropDistortion} training strategy was applied in a simple three-step way in this paper, and in future work it is of great worth to investigate self-adaptive \emph{DropDistortion}, where the distortion degree could be automatically adjusted.

\bibliographystyle{model2-names}

\end{document}